 \documentclass[final,5p,times,twocolumn,authoryear]{elsarticle}

\makeatletter
\def\ps@pprintTitle{%
 \let\@oddhead\@empty
 \let\@evenhead\@empty
 \let\@oddfoot\@empty
 \let\@evenfoot\@empty
}
\makeatother
\usepackage{amssymb}
\usepackage{lipsum}
\usepackage[utf8]{inputenc}
\usepackage{longtable}
\usepackage{blindtext,alltt}
\usepackage{booktabs} % for better looking tables
\usepackage{adjustbox} % for adjusting table width
\usepackage{rotating}
\usepackage{array}
\usepackage{graphicx} 
\usepackage{subcaption}
\usepackage{comment}
\usepackage{enumitem}
\usepackage{natbib}
\usepackage{subcaption}
\usepackage{amsmath}
\usepackage{caption}

\usepackage{natbib}
\renewcommand\harvardurl[1]{\textbf{URL:} \url{#1}}
\usepackage{hyperref}

\journal{High Energy Astrophysics}

\begin{document}

\begin{frontmatter}



\title{Predicting Mild Cognitive Impairment Using Naturalistic Driving and Trip Destination Modeling}


\author[me]{Souradeep Chattopadhyay}
\author[trans]{Guillermo Basulto-Elias}
\author[neuro]{Jun Ha Chang}
\author[neuro]{Matthew Rizzo}
\author[trans]{Shauna Hallmark}
\author[trans]{Anuj Sharma}
\author[me]{Soumik Sarkar}

\affiliation[me]{
    organization={Department of Mechanical Engineering, Iowa State University},
%    addressline={},
    city={Ames},
    state={IA},
%    postcode={},
    country={USA}
}

\affiliation[trans]{
    organization={Institute for Transportation, Iowa State University},
%    addressline={},
    city={Ames},
    state={IA},
%    postcode={},
    country={USA}
}

\affiliation[neuro]{
    organization={Department of Neurological Sciences, University of Nebraska Medical Center},
%    addressline={},
    city={Omaha},
    state={Nebraska},
%    postcode={},
    country={USA}
}

\begin{abstract}
Understanding the relationship between mild cognitive impairment (MCI) and driving behavior is essential for enhancing road safety, particularly among older adults. This study introduces a novel approach by incorporating specific trip destinations—such as home, work, medical appointments, social activities, and errands—using geohashing to analyze the driving habits of older drivers in Nebraska. We employed a two-fold methodology that combines data visualization with advanced machine learning models, including C5.0, Random Forest, and Support Vector Machines, to assess the effectiveness of these location-based variables in predicting cognitive impairment. Notably, the C5.0 model showed a robust and stable performance, achieving a median recall of 0.68, which indicates that our methodology accurately identifies cognitive impairment in drivers 68\% of the time. This emphasizes our model's capacity to reduce false negatives, a crucial factor given the profound implications of failing to identify impaired drivers. Our findings underscore the innovative use of life-space variables in understanding and predicting cognitive decline, offering avenues for early intervention and tailored support for affected individuals.

\end{abstract}



%\begin{keyword}
%% keywords here, in the form: keyword \sep keyword, up to a maximum of 6 keywords
%keyword 1 \sep keyword 2 \sep keyword 3 \sep keyword 4



%\end{keyword}

\end{frontmatter}

%\tableofcontents



\section{Introduction}
\label{introduction}

Addressing the impact of aging and cognitive decline on driving skills is becoming increasingly important, especially with the growing number of older drivers. As people age, they experience natural changes in memory, attention, and decision-making abilities, which can complicate their ability to plan and perform driving tasks effectively. These cognitive changes are often more significant in those diagnosed with Mild Cognitive Impairment (MCI) or Alzheimer's disease (AD). Many studies indicate that older adults with conditions like MCI and AD encounter significant challenges impacting their daily activities and raising serious concerns about road safety. For example, \citet{coxetal98} performed a simulation study on 29 outpatients with AD and found that drivers with AD made considerable driving errors, like often driving off the road, driving considerably slower than the posted speed limit, and having difficulty braking in stop zones. \citet{rizzoetal01} performed a simulation study with 30 participants and found that drivers with AD are more prone to car crashes at intersections compared to drivers without AD. \citet{ducheketal03} performed a study using 108 participants and found longitudinal evidence that driving performance decreases over time for participants with early-stage AD. A study by \citet{akinwuntan05} indicated that drivers with cognitive impairment, like dementia, face challenges with basic driving skills like lane changing, left turns, etc. \citet{foleyetal00} conducted a study using 152 participants and concluded that incident dementia is a significant cause of driving cessation, especially for older drivers of age 75 or beyond. \citet{lindaetal20} also conducted a study on 128 subjects, which indicated that drivers with AD type are more prone to crashes compared to non-demented drivers. These studies highlight the urgent need to develop studies that can shed light on the relationship between cognitive impairment and driving, especially for older adults, to ensure their safety and the safety of others sharing the road.

Data-driven approaches are highly effective in analyzing relationships between variables and can be particularly useful in exploring the connection between cognitive decline and naturalistic driving behavior.  Currently, a good amount of research has focused on using data-driven methods to unravel the complex relationship between cognitive decline and variables that reflect driving characteristics. \cite{hasanetal23} used a computer vision model to analyze driving patterns of older drivers with cognitive impairment across different road types and weather conditions. They found that weather-related driving responses can serve as an indicator of early signs of cognitive decline. In addition, \citet{hasanetal25} used naturalistic driving data from older drivers in Omaha to study how the cognitive decline of older drivers is affected, particularly under adverse weather conditions. Their analysis revealed that driving behavior can serve as an important indicator of cognitive decline, especially under adverse weather conditions. \citet{xuanetal21} performed a study to investigate the utility of 29 variables related to driving attributes like the number of trip chains, the total number of miles driven in a month, the number of left turns made in a month, the number of right turns made in a month, etc., and four demographic variables age, sex, ethnicity, and education to predict incident MCI and dementia in older adults. They trained several random forest classifiers using different driving characteristics and demographic variables. They found that these variables had a high predictive validity (88\%), implying that they are good predictors of MCI/dementia. \citet{bayatetal21} used random forest to investigate the relationship between 7 driving space behavior variables like average trip distance, total traveled distance, entropy, etc., and seven driving performance variables like average speed, average acceleration, overspeed, underspeed, etc., with preclinical AD. Using 139 participants aged 65 years and older, they trained their random forest model to differentiate between participants with preclinical AD from those without and achieved good accuracy, indicating that naturalistic driving behavior can serve as an efficient digital biomarker for early AD detection. \citet{derafshi2024impact} analyzed 246 older drivers, including 230 who were cognitively normal and 16 with incident cognitive impairment. They utilized geohashing techniques to identify common destinations and concluded that subjects with cognitive impairment used fewer distinct routes. 

Most previous studies have focused mainly on variables related to the drive space and characteristics of the vehicle. However, fewer studies have considered variables that reflect the driver's daily behavior. Metrics such as the number of trips to home, the number of trips taken to work, and the number of medical trips provide valuable insights into daily driving patterns and quality of life. These variables can help in understanding the effects of cognitive decline on naturalistic driving behavior. In this study, we examine the relationships between various driving behavior variables, such as the number of trips taken to home and work. We employ different machine learning techniques, including random forests and support vector machines, to investigate the effectiveness of these life-space variables in predicting the cognitive status of older drivers. Our approach also enables us to identify key factors that influence driving behavior in individuals with cognitive impairments as well as those aging normally.

\section{Description of the Dataset}
%\subsection{Data Collection}

%\subsubsection{Selection of Participants}
%\label{demographic}  

\begin{table}[]
    \centering
    \caption{Summary information of participants. Note that some participants did not provide demographic information, due to which no data was available for those participants. Also, some participants reported multiple employments.}
    \resizebox{\columnwidth}{!}{%
    \begin{tabular}{c p{3.7cm} c c}
    \hline\hline
     \textbf{Variable} & \textbf{Range or Counts}  & \textbf{Mean} & \textbf{Std. Dev.}\\
     \hline\hline
    Age (years) & Range: 65-92 & 76.57 & 6.13\\
    \hline
    Gender & Female: 76 & & \\
     & Male: 69 & &\\
     \hline
    Race & White: 134 & &\\
     & African American: 9 & &\\
     & American Indian: 1 & &\\
     & Asian: 1 & &\\
    \hline
    Employment & Employed (Full Time): 3 & &\\
     & Employed (Part Time): 25 & &\\
     & Homemaker: 2 & &\\
     & Volunteer: 16 & &\\
     & Retired: 127 & &\\
    \hline\hline 
    \end{tabular}
    }
    \label{tab:data_dem}
\end{table}

\subsection{Naturalistic Driving Assessment}
A total of 155 legally licensed local drivers were recruited from the Omaha, Nebraska area. Recruitment was done through fliers, local news, and talks with local senior organizations, see \cite{merickelRealworldRiskExposure2019}. All of the selected drivers consented to following institutional guidelines (IRB\#: 522-20-FB). All drivers selected for this study met Nebraska state driving license standards, including visual acuity better than or equal to 20/40 (corrected or uncorrected). Drivers with visual defects were permitted to participate if they met state license standards. Selected drivers exhibited a range of age-related dysfunctions typical of their age. Some demographic details of the 155 drivers are given in Table~\ref{tab:data_dem}. See \cite{chang2025day} for more details.

Naturalistic driving data were collected as part of a longitudinal study in which each driver participated for 3 months. The study focused on collecting information about driver functional abilities and driving behavior to assess driving risks with cognitive abilities. The personal vehicle of each selected driver was fitted with a `black box system' during the start of the study, which included custom-built sensors that monitored and recorded the driving behavior of every driver from on- to off-ignition. Each ignition-on and ignition-off instance was classified as a `drive' made by that particular driver. For every drive, the onboard accelerometer, video, and vehicle sensor collected different kinds of data like driving video, speed, acceleration, etc., every second during that particular drive. In addition, a global positioning system (GPS) onboard the black box recorded the latitude and longitude of the driver's position every second from ignition on (start of drive) to ignition off (end of drive).

\subsubsection{Laboratory Assessments}
The driver laboratory assessment consisted of both a socioeconomic survey and a clinical diagnosis to determine their cognitive status.

As a part of the driving location survey, drivers were asked to provide the following information:
\begin{enumerate}[label=(\roman*)]
    \item Home address.
    \item Workplace name and address.
    \item Name and address of regularly visited places, such as:
    \begin{enumerate}
        \item  Locations of daily errands (e.g., gas, groceries, prescriptions).

        \item Locations of social activities (for example, religion, exercise, restaurants/clubs).

         \item Locations they visit for Medical appointments or health care needs.   
    \end{enumerate}
\end{enumerate}
 Based on the survey, the locations most frequently visited by the  drivers were grouped into nine main categories: `home,' `work,' `groceries,' `gas,' `prescriptions,' `social,' `exercise,' `religion,' and `doctor.’ In addition to these primary categories, drivers also reported other regular destinations, such as visits to family or friends, which were categorized as 'other'.

In addition to a demographic survey, each selected driver completed a driver behavior survey (primary driving environment, driving experience, and driving frequency) and health (medication usage, diagnostic history). Selected drivers were classified as cognitively unimpaired (CU) and mild cognitive impairment (MCI) groups based on the 2018 NIA-AA research criteria for syndromal staging of the cognitive continuum. For each driver, neurological and neuropsychological assessments were performed as per the National Alzheimer's Coordinating Center (NACC) Unified Data Set (UDS) version. They came to a consensus on cognitive staging. Further, to assess cognitive decline, the Montreal Cognitive Assessment (MoCA) and neuropsychological test battery were administered to every selected driver in five cognitive domains: memory (Benson Recall, Craft Story Delay – Verbatim, HVLT Recognition, and HVLT Delay), language (category fluency tests for animals and vegetables, as well as Multilingual Naming Test [MINT)), visuospatial skills (Benson Copy and WAIS-III Block Design Test), executive function (Trail Making Test Part B and Verbal Fluency Tests for letters F and L), and attention (Trail Making Test Part A, digit span forward for the number of correct strings recalled, and digit span backward for the number of correct strings recalled). All test scores were adjusted for z-scores using age, sex, and education. Subsequently, Jak / Bondi (at least two cognitive tests per domain, $>$ 1 SD below the norms) were used to identify drivers who fall into the MCI category. Of the 155 subjects selected for our analysis, 61 were classified as MCI/AD and 94 as CU. Henceforth, we refer to this variable as the 'cognitive ability' (CA) of the driver. 

\subsection{Data Preprocessing}
Before computing our naturalistic driving variables, we performed an extensive preprocessing of the data to ensure the dataset is free of unwanted characteristics that might not reflect the naturalistic driving behavior of the drivers. The steps taken to preprocess the data are as follows:

\begin{enumerate}

    \item Drives that had missing or unrecorded `start/end of drive' coordinates were removed, as these were classified as incomplete drives. This led to the removal of 11,578 drives. 

    \item Drives in which the driver did not personally operate the vehicle, but instead permitted someone else to drive, were excluded. Additionally, drives made to the University of Nebraska Medical Center for routine maintenance of the black box devices were also removed from consideration. In total, 395 such drives were excluded.

    \item For each driver, we removed drives where the distance between the start and end points was less than 0.2 miles. This eliminated very short drives where the vehicle was turned on but did not actually move, or instances where it was relocated within the same area. A total of 3,839 such drives were excluded.
\end{enumerate}

Also, since this study primarily focused on drivers in Nebraska, drives ending outside the state boundaries of Nebraska were excluded. We identified and removed these drives by checking if the end drive latitude and longitude coordinates fell outside Nebraska's geographic borders, given by the state's bounding box. This process also led to the exclusion of all drivers for two drivers who were based in Iowa. At the end of our preprocessing, we were left with 153 drivers and a total of 29,192 end drives. 

We now discuss our adopted approach for the computation and analysis of the variables used in our study. To ensure a robust analysis, we have taken a 2-fold approach, details of which are discussed in the next section.
\section{Methodology}
This section begins with a detailed description of our adopted approaches for the computation of life-space variables, followed by which is a discussion of the approaches taken to explore the relationship between cognitive ability and computed variables.
\subsection{Computation of Driving Behavior Variables}
\label{sec:lifespace}
To efficiently capture the naturalistic driving behavior of the drivers, we computed variables for each driver that can capture the daily driving behavior of the drivers. For each driver, our variables indicate the driver's driving behavior based on the nine known location categories. Our six chosen variables are as follows:

\begin{enumerate}[label=(\roman*)]
    \item \textbf{Home trips}: The number of trips that ended near `home' location of the driver.

    \item \textbf{Work trips}: The number of trips that ended near `work' location of the driver.

    \item \textbf{Errand trips}: The number of trips that ended near locations visited by the driver for daily errands. Of the ten categories, trips ending near `groceries,' `gas,' and `prescriptions' locations constitute an errand trip.

    \item \textbf{Medical trips}: The number of trips that ended near locations visited by the driver for medical visits. Of the ten categories, trips ending near `doctor' location constitute a medical trip.

    \item \textbf{Social trips}: The number of trips that ended near locations visited by the driver for social activities. Of the ten categories, trips ending near the `social', `exercise', and `religion' locations constitute a social trip.

    \item \textbf{Unknown trips}: The number of trips that ended at locations different from the ten main categories.
\end{enumerate}

We call these variables \textit{life-space variables} since these variables not only reflect the daily driving behavior, but also give an indication of the quality of life of the drivers.

To better understand naturalistic driving behaviors, we calculated each of the six life-space variables separately for weekdays and weekends. This distinction was crucial because drivers with Mild Cognitive Impairment (MCI) or AD tend to exhibit different driving behaviors on weekdays compared to weekends, and this should be accounted for in our life-space variables.

Given the data, one might feel a straightforward approach for computing these life-space variables would be by comparing the end drive coordinate of a given trip with the coordinates of the ten categories, a match of which would indicate a trip to that location. But direct comparison of end drive coordinates with the ten location coordinates can prove extremely challenging, mainly because latitudes and longitudes differ significantly even across minimal distances. For example, the latitude and longitude of a particular building can be considerably different from the latitude and longitude of a location in the parking lot of the building, as a result of which the coordinates of a drive ending at a parking lot of the building will be different from the coordinates of the building.

Geohashing converts latitudes and longitudes into a compact, Base32-encoded string. It works by recursively subdividing the Earth's surface into a grid of bounding boxes, each represented by a string. The length of the geohash string determines its precision: longer strings denote smaller bounding boxes, allowing more accurate representations of specific locations. A key feature of geohashes is that geographically close points share common geohash prefixes. This means two nearby locations, such as a building and a location in its parking lot, will typically have identical geohashes of the same length. As the geohash string length increases, the bounding box shrinks, providing greater spatial resolution. For more details on computing geohashes, see \citet{suwardietal15}. 

\subsubsection{Computation of Life-Space Variables}
As discussed in the previous section, using geohash to compute life-space variables effectively addresses the challenge of direct comparison, which makes it a perfect tool for our use case. 

Suppose a driver, denoted by $D_e$, $e = 1,2,\ldots, 153$ has $n_e$ end drives. Let $d_{eiw}$, $i = 1,2,\ldots,n_e$, denote the $i$th drive for the $e$th driver, where $w = 1$ if the drive took place was on a weekday and zero if it took place on a weekend. Let $G_{d_{eiw}}$, $i=1,2,\ldots,n_e$ denote the geohash of the $i$th drive for the $e$th driver. Also, let the geohashes of the ten known locations for $e$-th driver be $G^{home}_{D_e}$, $G^{work}_{D_e}$, $G^{doctor}_{D_e}$, $G^{groceries}_{D_e}$, $G^{prescriptions}_{D_e}$, $G^{gas}_{D_e}$, $G^{social}_{D_e}$, $G^{religion}_{D_e}$, $G^{exercise}_{D_e}$ and $G^{other}_{D_e}$. Now define
\begin{align}
    T^l_{eiw} &= 
    \begin{cases}
        1, & \text{if } G_{d_{eiw}} = G^{l}_{D_e}, \\
        0, & \text{otherwise,}
    \end{cases}
\end{align}
 where $l$ = \{`home', `work', `doctor', `groceries', `prescriptions', `gas', `social', `religion', `exercise', `other'\}, $e = 1,2,\ldots153$, $i = 1,2,\ldots,n_e$, and $w = {0, 1}$. Then the number of drives to the $l$th for $e$th driver $ND_{lew}$ can be computed as

 \begin{equation}
     ND_{lew} = \sum_{i=1}^{n_e} T^l_{eiw}.
\end{equation}

Upon obtaining $ND_{lew}$ for every value of $l$ and $e$, the life-space variables for $D_e$ can be computed as:

\begin{enumerate}[label=(\roman*)]
    \item \textbf{Home trips (Weekday)} = $ND_{lew}$, $l$ = `home', $w = 1$.
    
     \item \textbf{Home trips (Weekend)} = $ND_{lew}$, $l$ = `home', $w = 0$.

     \item \textbf{Work trips (Weekday)} = $ND_{lew}$, $l$ = `work', $w = 1$.
     
     \item \textbf{Work trips (Weekend)} = $ND_{lew}$, $l$ = `work', $w = 0$.

     \item \textbf{Errand trips (Weekday)} = $\sum_\mathcal{K} ND_{lew}$, $\mathcal{K}$ = \{`groceries', `prescriptions', `gas'\}, $w = 1$.

     \item \textbf{Errand trips (Weekend)} = $\sum_\mathcal{K} ND_{lew}$, $\mathcal{K}$ = \{`groceries', `prescriptions', `gas'\}, $w = 0$.

     \item \textbf{Medical trips (Weekday)} = $\sum_\mathcal{K} ND_{lew}$, $\mathcal{K}$ = \{`doctor'\}, $w = 1$.

     \item \textbf{Medical trips (Weekend)} = $\sum_\mathcal{K} ND_{lew}$, $\mathcal{K}$ = \{`doctor'\}, $w = 0$.

     \item \textbf{Social trips (Weekday)} = $\sum_\mathcal{K} ND_{lew}$, $\mathcal{K}$ = \{`social', `exercise', `religion'\}, $w = 1$.

     \item \textbf{Social trips (Weekend)} = $\sum_\mathcal{K} ND_{lew}$, $\mathcal{K}$ = \{`social', `exercise', `religion'\}, $w = 0$.

     \item \textbf{Unknown trips (weekday)} = $\sum_\mathcal{K} ND_{lew}$, $\mathcal{K} \neq$ \{`home', `work', `doctor', `groceries', `prescriptions', `gas', `social', `religion', `exercise', `other'\}, $w = 1$.

     \item \textbf{Unknown trips (weekend)} = $\sum_\mathcal{K} ND_{lew}$, $\mathcal{K} \neq$ \{`home', `work', `doctor', `groceries', `prescriptions', `gas', `social', `religion', `exercise', `other'\}, $w = 0$.
\end{enumerate}

After computing the life-space variables, two bottlenecks remained: (1) characterization of trips ending at `other' and (2) characterization of `Unknown trips.'

Locations labeled as `other' are trips not among the nine central locations but were still potentially for purposes such as `errand,' `social,' or `medical.' Hence, further investigation of these drives was essential to ensure better accuracy of the life-space variables. To better understand the nature of these `other' category trips, the following steps were carried out:

\begin{enumerate}[label=(\roman*)]
    \item Drives labeled as `other' were segregated for each driver.
    \item The center coordinates of the geohash bounding box for each end drive were calculated.
    \item These coordinates were fed into Google Maps to identify the end drive location.
    \item Depending on the proximity to landmarks, they were relabeled as:
    \begin{enumerate}[label=(\alph*)]
        \item `Social trip' if the closest landmark were clubs, restaurants, parks, etc.)
        \item `Errand trip' if the closest landmark were grocery stores, gas stations, etc.
        \item `Medical trip' if the closest landmark were hospitals, medical centers, etc.
        \item `Unknown trip' if the closest landmark cannot be categorized as a social, errand, or medical trip.
        \item  The life-space variables were updated based on the relabeled `other' trips.
    \end{enumerate}
\end{enumerate}

At the end of our labeling process, a large number of `Unknown trips' labels remained, which were relabeled following the same method as that for the `other' location category. This was important since even though these trips were not labeled during data collection, they could still be essential for life-space variables.

\textbf{Removal of Multi-label trips}: Some trips were labeled with multiple categories because their geohash had various matches with the ten known location categories. For example, geohashes of end drives of some trips matched with both `work' and `social,' hence becoming a multi-label trip. This typically happened when work and social locations were close, making their geohashes similar. Since it was impossible to determine the actual end points of these drives, all multi-label trips were excluded from the analysis.  

Finally, the life-space variables were divided by the number of days the driver participated in the study and then multiplied by 100 to determine the average number of drives per 100 days. This standardized data was then used to analyze the relationship between cognitive impairment and life-space variables.

Our adopted approach consists of two strategies: (1) data exploration and visual analysis, and (2) model-based analysis of life-space variables. Details are provided in the following sections.

\subsection{Data Exploration and Visual Analysis of Life-Space Variables}
Before performing any modeling on the data, it is often necessary to perform some kind of forensic investigation using simple descriptive and visualization methods, since these simple methods usually provide essential insight into the nature of the data and help decide the model choices. Keeping this in mind, we conducted a two-step analysis using simple descriptive measures, which are as follows:

\begin{enumerate}[label=(\roman*)]
    \item \textbf{Descriptive Statistics Analysis:}
    \begin{itemize}
        \item Computed summary statistics such as mean and standard deviation for each life-space variable to understand their central tendency and dispersion characteristics.
    \end{itemize}

    \item \textbf{Visual Exploration:}
    \begin{itemize}
        \item Created radial plots to visually represent the distribution and patterns of life-space variables for a comprehensive data view.
    \end{itemize}
\end{enumerate}

This strategic analysis helps shed light on the relationship between life-space variables and cognitive ability. Also, it helps understand the importance of each of these life-space variables relative to each other. 

\subsection{Model-Based Analysis of Life-Space Variables}
This section discusses our adopted modeling approach to analyze the relationship between life-space variables and cognitive ability. An important thing to note here is that even though our sample size is conservative (153 samples) due to data collection limitations, our sample size is bigger compared to previous studies like \citet{bayatetal21}. However, care must be taken to ensure that the conclusions drawn from the fitted models are robust and stable due to the conservative sample size. A common practice during any modeling is randomly splitting the dataset into training and test sets, followed by which the model is created using the training set and evaluated using a test set. When the number of data points is small, this approach can bring forth many challenges, including instability in model performance. This typically implies that the model will demonstrate varying performance for different train-test set combinations, making informed decisions using the fitted model will become challenging. Thus, care must be taken to ensure that a model fitted on small datasets exhibits robustness and stability and doesn't provide unreliable inferences.

Since our goal here is to understand the relationship between the twelve life-space variables and cognitive ability, our first step involved finding a robust model that effectively captures the relationship between cognitive ability and the twelve life-space variables and exhibits stability across different train-test sets. To achieve this, we adopted a resampling procedure by selecting different train-test sets from the data and fitting three popular classification models, Random Forest (RF), Support Vector Machines (SVM), and C5.0. 

C5.0~\citep{franketal98} is a popular algorithm that builds a single optimized classification/regression tree by using boosting techniques, whereas RF~\citep{breiman201} builds multiple decision trees and then combines predictions from numerous trees using ensembling methods. RFs have proven to be a robust method for complex datasets with many variables, but can sometimes be challenging concerning the interpretability of results, especially for smaller datasets. C5.0, on the other hand, is easier to interpret since it builds a single tree and has been shown to work well with small datasets. However, compared to RF, C5.0 is more prone to overfitting, sometimes making it harder to fit on complex datasets. SVM~\citep{hearstetal98} is another popular algorithm, especially for binary classification problems. SVM tries to find an optimal line/hyperplane that can best separate the different classes in the data. However, SVM typically works best for linearly separable classes, beyond which it can become challenging and computationally expensive for large datasets. 

Since each of the three algorithms has advantages and disadvantages, we fitted all three to determine the best algorithm suitable for our dataset. Details of our approach as as follows:

\begin{enumerate}
    \item Let $\mathcal{X}$ denote the life-space variables and $\mathcal{Y}$ denote cognitive ability.

    \item  ($\mathcal{X}$, $\mathcal{Y}$) was sampled 1000 times to generate 1000 train-test sets. Let the train sets be denoted by ($\mathcal{X}_{Tr_i}$, $\mathcal{Y}_{Tr_i}$) and the test sets by ($\mathcal{X}_{Te_i}$, $\mathcal{Y}_{Te_i}$), $i=1,2,\ldots,1000$. For each sampling case, we used 80 \% for training and the rest for testing.

    \item For each ($\mathcal{X}_{Tr_i}$, $\mathcal{Y}_{Tr_i}$) SVM, RF, and C5.0 with 10-fold cross validation for hyperparameter tuning were fitted. Let the models for $i$-th  train test set be ${\mathcal{M}_{svm}}_i$, ${\mathcal{M}_{rf}}_i$, and ${\mathcal{M}_{c50}}_i$,  $i=1,2,\ldots,1000$ for SVM, RF, and C5.0, respectively. 

    \item Using ${\mathcal{M}_{svm}}_i$, ${\mathcal{M}_{rf}}_i$, ${\mathcal{M}_{c50}}_i$ and $\mathcal{X}_{Te_i}$ predictions are obtained for $\mathcal{Y}_{Te_i}$, $i=1,2,\ldots,1000$. Let the predicted values for SVM, RF, and C50 be denoted as $\hat{\mathcal{Y}_{svm}}_{Te_i}$, $\hat{\mathcal{Y}_{rf}}_{Te_i}$, and $\hat{\mathcal{Y}_{c50}}_{Te_i}$, respectively.

    \item Using ($\mathcal{Y}_{Te_i}$, $\hat{\mathcal{Y}_{x}}_{Te_i}$) the accuracy is obtained as $\mathcal{A}_{x} (\mathcal{Y}_{Te_i}, \hat{\mathcal{Y}_{x}}_{Te_i})$ = (Number of correct predictions from $x$)/(Total number of samples in $\mathcal{Y}_{Te_i}$), $i=1,2,\ldots,1000$, $x = \{svm, rf, c50\}$.

    \item The first, second, and third quartiles of the accuracies for each model are investigated to determine the model with the highest accuracy and best stability.
\end{enumerate}

Once the best model was determined, we conducted a further analysis using the selected model to determine how often drivers got misclassified with respect to CA for the 1000 train-test samples. Additional analysis was also carried out using MoCA and COGSTAT scores of the drivers in order to investigate the cognitive ability of the drivers who were more prone to misclassification. MoCA is an efficient screening tool used to detect early signs of MCI and AD through evaluation of memory, attention, language, executive function, and visuospatial abilities where whereas COGSTAT is a measure of cognitive function calculated as a summation of standardized T-scores across eight different cognitive tests from the cognitive test battery~\citealp{dawsonetal10}.
The remainder of this paper discusses the results obtained using both model-free and model-based approaches and also analyzes the implications of the obtained results from a medical point of view.

\section{Results and Discussions}
We begin our discussion with the results of our data exploration and visualization, followed by which we proceed to discuss of the results of the model-based approach.  
\subsection{Data Exploration and Visualization of Life-Space Variables}
\begin{table*}
\hspace*{-2.5cm}
    \centering
    \caption{Summary measures of the life-space variables for the two cognitive ability categories.}
    \begin{tabular}{l|rrr}
        \hline\hline
        \textbf{Cognitive Ability} & \textbf{Life-space Variable} & \textbf{Mean} & \textbf{Standard Deviation (SD)} \\
        \hline\hline
        MCI/AD & Home trip (Weekday) & 80.61 & 48.87\\
        & Home trips(Weekend) & 24.96 & 18.58\\
        & Work trips (Weekday) & 2.29 & 9.42 \\
        & Work trips (Weekend) & 0.59 & 2.64 \\
        & Errand trips (Weekday) & 79.61 & 38.95 \\
        & Errand trips (Weekend) & 20.44 & 14.82 \\
        & Medical trips (Weekday) & 12.26 & 12.92 \\
        & Medical Trips (Weekend) & 2.07 & 5.42\\
        & Social trips (Weekday) & 52.30 & 35.89 \\
        & Social trips (Weekend) & 15.82 & 14.32 \\
        & Unknown trips (Weekday) & 27.58 & 40.87\\
        & Unknown trips (Weekend) & 8.69 & 12.8 \\
        \bottomrule
        Cognitively Unimpaired & Home trip (Weekday) & 74.86 & 43.51\\
        & Home trips(Weekend) & 22.18 & 15.87\\
        & Work trips (Weekday) & 2.13 & 5.57 \\
        & Work trips (Weekend) & 0.58 & 2.91 \\
        & Errand trips (Weekday) & 81.02 & 39.43 \\
        & Errand trips (Weekend) & 20.89 & 19.91 \\
        & Medical trips (Weekday) & 9.05 & 7.77 \\
        & Medical trips (Weekend) & 0.75 & 1.50\\
        & Social trips (Weekday) & 54.22 & 29.90 \\
        & Social trips (Weekend) & 17.48 & 13.97 \\
        & Unknown trips (Weekday) & 28.52 & 36.63\\
        & Unknown trips (Weekend) & 9.71 & 12.52 \\
        \hline\hline
    \end{tabular}
    \label{tab:lifespace_summary}
\end{table*}

\begin{comment}
\begin{table}[ht]
\centering
\caption{Biserial Correlation Between cognitive impairment and life-space variables} 
\label{tab:biserial_correlation}
\begin{tabular}{lr}
  \hline\hline
Life-space Variable & Biserial Correlation \\ 
  \midrule
Errand trips (Weekday) & 0.006 \\ 
  Home trips (Weekday) & -0.06 \\ 
  Medical trips (Weekday) & -0.12 \\ 
  Social trips (Weekday) & 0.04 \\ 
  Errand trips (Weekend) & 0.02 \\ 
  Home trips (Weekend) & -0.05 \\ 
  Unknown trips (Weekend) & 0.07 \\ 
  Social trips (Weekend) & 0.08 \\ 
  Unknown trips (Weekday) & 0.05 \\ 
  Work trips (Weekday) & 0.02 \\ 
  Medical trips (Weekend) & -0.13 \\ 
  Work trips (Weekend) & 0.02 \\ 
   \bottomrule
\end{tabular}
\end{table}
%\begin{center}
\end{comment}
\begin{figure*}
  \centering
    \begin{subfigure}{0.45\linewidth}
    \centering
    \includegraphics[width=\textwidth]{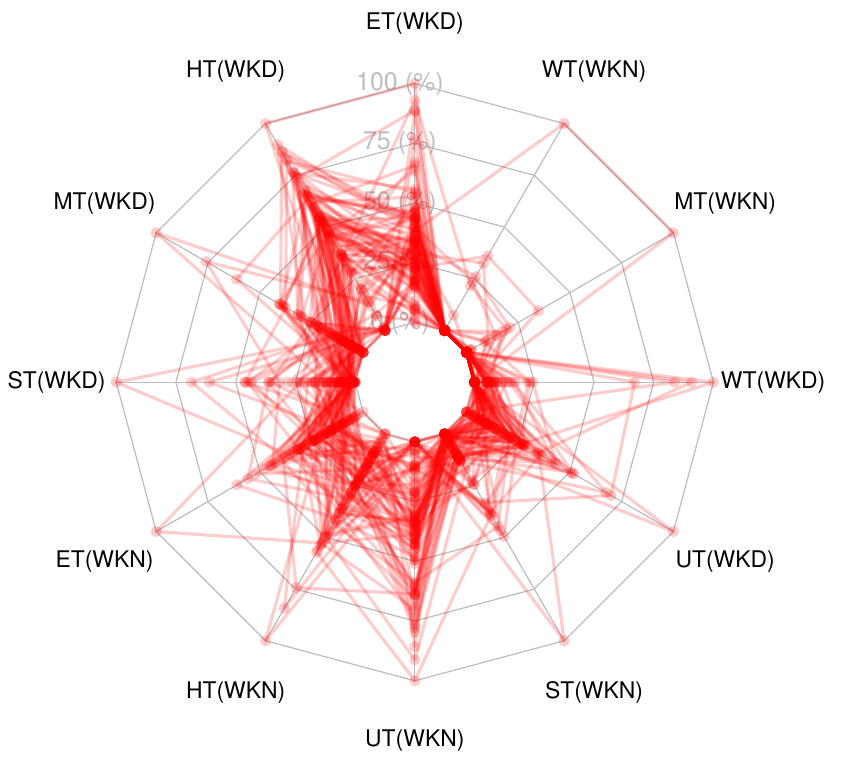}
    \caption{}
    \end{subfigure}
    \hspace{0.5cm} % Adjust the horizontal space between the two subfigures
    \begin{subfigure}{0.45\linewidth}
    \centering
    \includegraphics[width=\textwidth]{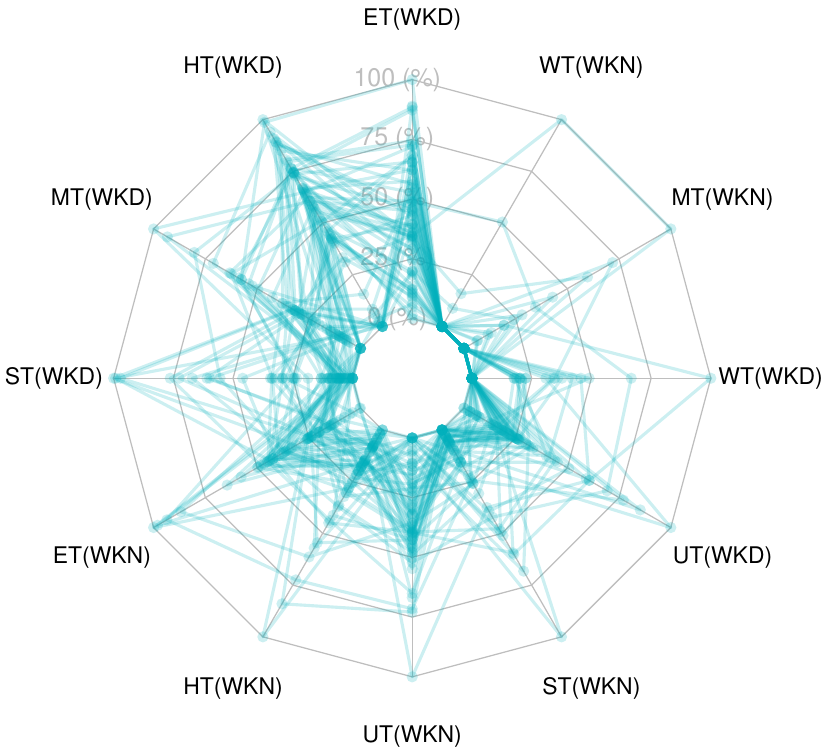}
    \caption{}
    \end{subfigure}
    \caption{Radial plots of life-space variables for (a) drivers with MCI/AD and (b) cognitively unimpaired drivers. Here, HT refers to home trips, ET refers to errand trips, ST refers to social trips, MT refers to medical trips, WKD refers to weekdays, and WKN refers to weekends.}
    \label{fig:radian}
\end{figure*}
 
%\end{center}

Table~\ref{tab:lifespace_summary} presents the summary measures of each standardized life-space variable. A multivariate visualization of the life-space variables for both classes of cognitive ability is also presented in Figure~\ref{fig:radian}. Both classes show high mean values for home trips during weekdays, which indicates that drivers of both categories made frequent trips home during weekdays. However, drivers with MCI/AD have a higher SD compared to cognitively unimpaired drivers, indicating more variability in their driving patterns. The mean is low for both home trips during weekends, which means that their driving behavior with respect to home trips does not vary much for both drivers with MCI/AD and cognitively unimpaired drivers. 

Both groups of cognitive ability made very low work trips during weekdays, while during weekends, they made almost no work trips. This is expected since our study cohort consists only of drivers aged greater than 65. Both categories of drivers made regular trips for errands during both weekdays and weekends, as evidenced by the similar SDs across both groups. However, average daily trips for errands were significantly higher during the weekdays compared to weekends. 

With respect to medical trips, some important differences exist between the two groups. Drivers with MCI/AD made more average daily medical visits during weekdays compared to cognitively unimpaired drivers. This is also true for weekends, where cognitively unimpaired drivers made very few trips, while drivers with MCI/AD made significantly more daily average medical trips during the weekend compared to cognitively unimpaired drivers. The frequency of social trips is moderate for both groups on weekdays, but slightly higher for the cognitively unimpaired group compared to MCI/AD. This indicates that drivers having MCI/AD showed a little less participation in social events compared to cognitively unimpaired drivers during weekdays. The MCI/AD group has a slightly lower mean for unknown trips during weekdays compared to cognitively unimpaired, but both groups have similar variability, suggesting that both groups show a range of driving patterns for trips to locations other than the ten known locations.

Apart from the summary measures, the radial plots also exhibit differences in distribution between drivers with MCI/AD and cognitively unimpaired drivers. The radial plots indicate that medical trips and social trips show considerable differences across the two categories. Apart from this, unknown trips also contribute to establishing differences in distribution between MCI/AD. This points to the fact that driving behavior outside common locations serves as an important indicator of the cognitive ability of drivers. However, it is important to note that summary measures and visualization only provide basic ideas about the nature of the relationship between cognitive ability and life-space variables, but cannot capture any complex forms of relationship between them. Hence, to investigate any other form of relationships, we conducted our model-based approach, the results of which are discussed in the next section.

\subsection{Model-Based Approach}
This section is subdivided into three parts. In the first part, we discuss our analysis of the best model obtained for understanding the relationship between cognitive ability and life-space variables. The second part presents a detailed analysis of the classification results obtained using the best-fitted model using MoCA and COGSTAT scores. The final part discusses the importance of the different life-space variables in determining the cognitive ability of the drivers. Note that for our model-based analysis, we excluded one particular driver, which had just a single medical trip and no other trips, and two drivers who had missing COGSTAT scores, thus making our total sample size 150.
\subsection{Selection of robust model}
\begin{comment}
\begin{table*}[ht]
\centering
\caption{Summary statistics (first quartle (Q1), Median, third quartile (Q3 and standard deviation(SD)) for Support Vector Machine (SVM), Random Forest (RF) and C5.0}
\begin{tabular}{lcccc}
\hline\hline
\textbf{Model} & \textbf{Q1} & \textbf{Median} & \textbf{Q3} & \textbf{SD} \\
\hline\hline
C5.0   & 0.59 & 0.59 & 0.59 & 0.03 \\
RF    & 0.47 & 0.53 & 0.59 & 0.07\\
SVM   & 0.56 & 0.59 & 0.59 & 0.04\\
\hline
\end{tabular}
\label{tab:best_mod}
\end{table*}
\end{comment}

\begin{table*}[ht]
\centering
\caption{From left to right, (a) accuracy for  Support Vector Machine (SVM), Random Forest (RF), and C5.0, and (b) performance metrics for C5.0.}
\begin{minipage}{0.45\textwidth}
    \centering
    \begin{tabular}{lcccc}
    \hline\hline
    \textbf{Model} & \textbf{Q1} & \textbf{Median} & \textbf{Q3} & \textbf{SD} \\
    \hline
    C5.0 & 0.59 & 0.59 & 0.59 & 0.04 \\
    RF   & 0.46 & 0.50 & 0.54 & 0.05 \\
    SVM  & 0.55 & 0.59 & 0.59 & 0.04 \\
    \hline\hline
    \end{tabular}
    \subcaption{Summary statistics for accuracy (first quartile (Q1), median, third quartile (Q3), and standard deviation (SD)).}
    \label{tab:acc_sum}
\end{minipage}%
\hspace{0.05\textwidth} % Adjusts the space between the tables
\begin{minipage}{0.45\textwidth}
    \centering
    \begin{tabular}{lccc}
    \hline\hline
    \textbf{Metric} & \textbf{Q1} & \textbf{Median} & \textbf{Q3} \\
    \hline
    Precision & 0.54 & 0.56 & 0.59 \\
    Recall   & 0.62 & 0.68 & 0.75 \\
    F1 Score  & 0.58 & 0.62 & 0.66 \\
    \hline\hline
    \end{tabular}
    \subcaption{Q1, Q2 and Q3 of Precision, Recall and F1 scores for C5.0.}
    \label{tab:prec_sum}
\end{minipage}
\label{tab:sum_and_prec}
\end{table*}

This section discusses the results of our resampling approach used to determine the best model for characterizing the relationship between life-space variables and cognitive ability. Our choice of the best model was primarily based on two characteristics: (1) how accurate the model is in predicting cognitive ability and (2) how stable the model is in terms of accuracy when fitted for different train-test sets.

Table~\ref{tab:acc_sum} presents the first, second, and third quartiles of the 1000 accuracies obtained from the 1000 sampled train-test pairs. It is evident that C5.0 performs better both with respect to accuracy and stability compared to SVM and RF. The fact that all three quartiles for C5.0 are the same indicates that the accuracy did not change significantly for the 1000 different train-test pairs compared to RF and SVM. It is important to note that SVM also performs well, and its accuracies are very similar compared to those of C5.0. But since C5.0 exhibits more stability with respect to the quartiles, we select C5.0 as our model of choice for further analysis

In addition to the accuracy, we also computed the precision, recall, and F1 scores for C5.0 (with MCI/AD as a positive class), which are given in Table~\ref {tab:prec_sum}. The precision has a median of 0.56, which indicates that the model is able to predict MCI/AD in about 56\% of the cases. Also, a median recall value of 0.68 suggests that, on average, the model correctly identifies actual MCI/AD cases 68\% of the time. Apart from this, the third quartile of recall scores is 0.75, which indicates that C5.0 correct identified MCI/AD cases 75\% for a significant number of train-test pairs. Our results are comparable to those of \citet{dietal21}, who performed a similar study of predicting incident MCI using driving behavior. They fitted several models using demographic covariates like age, sex, education, and driving covariates like 'total number of miles driven in month' and 'percent of trips traveled in month within 15 miles of home. Using driving covariates, their model achieved a recall of 0.56, whereas our model achieved a recall of 0.68. This highlights the potential of the variables included in our model, which have not been used in similar models to the best of our knowledge. Typically, in screening studies like ours, it is desirable to have high recall rates since high recall rates are an indication that the model is capable of correctly identifying the positive cases, which here is MCI/AD. Overall, given the sample size, the model is able to perform well and thus is suitable for studying the importance of the life-space variables in predicting CA. Also, the results indicate that with more data, the model can predict MCI/AD cases with good accuracy. 

In the next sections, we do a more detailed analysis of the model by computing the misclassification rates of the model and also study the MoCA and COGSTAT scores to gain a better understanding of the cognitive status of the drivers.

\subsection{Analysis of MoCA and COGSTAT Scores}
\begin{comment}
\begin{figure}[htbp]
    \centering
    \includegraphics[width=0.8\textwidth]{moca_vs_misclassify.pdf}
    \caption{MOCA scores vs misclassification probabilities for all 152 drivers. Red indicates drivers drivers with MCI/AD.}
    \label{fig:mocavsmis}  % Reference label for this figure
\end{figure}
\end{comment}
\begin{figure*}[htbp]
    \centering
    \mbox{
        \includegraphics[width=0.5\textwidth]{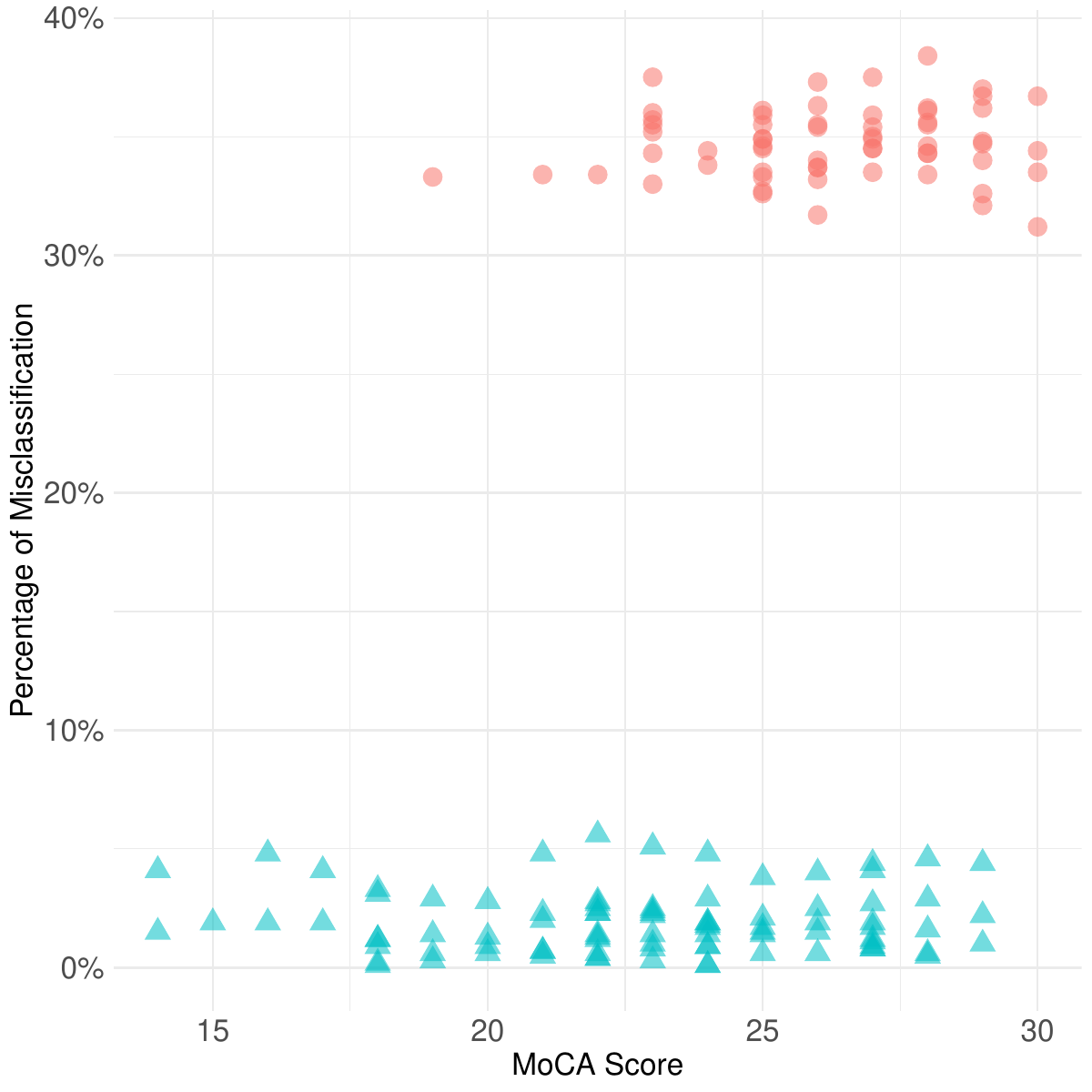}
        \hspace{0.02\textwidth} 
        \includegraphics[width=0.5\textwidth]{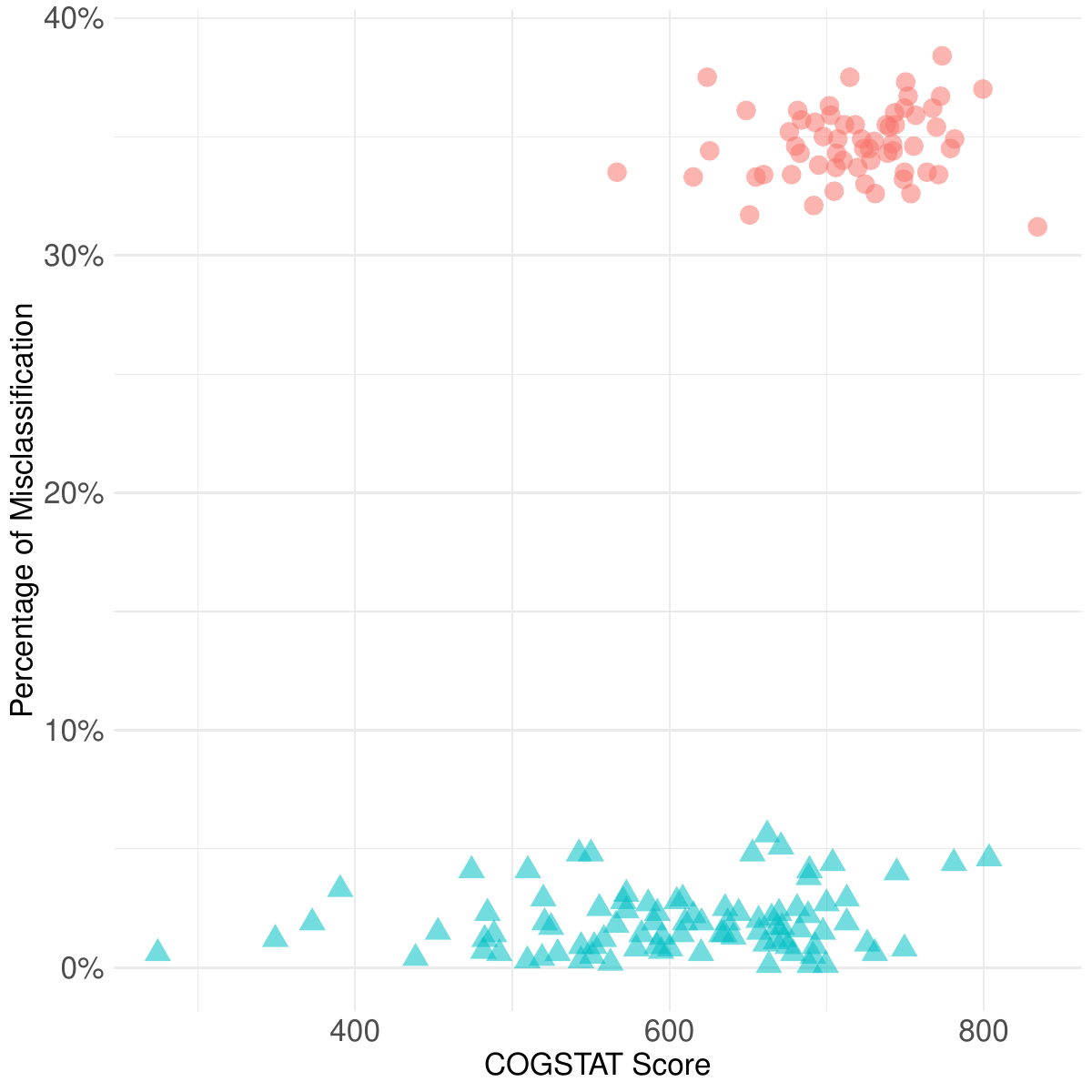}  
    }
    \caption{From left, MoCA scores vs misclassification probabilities for all 150 drivers. Red indicates drivers with MCI/AD, COGSTAT scores vs misclassification probabilities for all 150 drivers. Red indicates drivers with MCI/AD}
    \label{fig:comparison_moca_cog}
\end{figure*}

In this section, we discuss in greater detail the results of sampling analysis using C5.0 to understand the accuracy of C5.0 in predicting cognitive ability from life-space variables.

To gain a better understanding of the efficiency of C5.0, we computed the percentage of misclassification for each of the 150 drivers. For the $i$-th driver, $i=1,2,\ldots
150$, the percentage of misclassification was computed as
\begin{enumerate}[label=(\roman*)]
    \item Let $n_{test}$ be the number of times the $i$-th driver appeared in a test set out of 1000 resamples.

    \item Let $n_{correct}$ be the number of times out of $n_{test}$ when the $i$-th driver was misclassified.

    \item The percentage of misclassification for the $i$-th driver was computed as $100 \times (n_{correct}/n_{test})$.
\end{enumerate}
 The misclassification percentages give us a better sense of understanding the effectiveness of C5.0 as a modeling choice. In our case, most drivers showed a low misclassification percentage ($<$10\%) while a few of them exhibited a misclassification percentage in the range of 30-40\%. A good number of drivers exhibited misclassification of less than 1\%. This indicates that C5.0 indeed performed a good job of correctly identifying the CA of the drivers based on variable importance.

 To further investigate the cognitive states of the drivers, we plotted the MoCA and COGSTAT scores of the drivers versus their percentage of misclassification, which is shown in Figure~\ref {fig:comparison_moca_cog}. It is evident that drivers with high misclassification rates ($>$ 30\%) show higher MoCA and COGSTAT scores compared to drivers with lower misclassification rates. Drivers with high misclassification rates mostly fall in the cognitively unimpaired category, indicating that even though our model predicts MCI/AD cases well, there is much room for improvement for the cognitively unimpaired category. We believe this can be achieved by obtaining more data for each of the two categories to ensure the model has sufficient data to learn about both cases efficiently.

 Since our model showed good potential in identifying drivers with MCI/AD, we analyzed the importance of the life-space variables in identifying patients with MCI/AD.

\subsection{Analysis of Life-Space Variable Importances}

To calculate the importance of each life-space variable, we used the fitted C5.0 model to determine the efficiency of each of the variables in segregating MCI/AD from cognitively unimpaired drivers. The C5.0 algorithm computes variable importance by evaluating the effectiveness of each variable in reducing prediction errors while making splits. Variables that contribute more to reducing prediction errors by being frequently selected across splits receive higher importance scores. More details of this method can be found in \cite{franketal98}. 

Since we fitted 1000 different models on different train-test sets, we computed the importance of each life-space variable for all 1000 cases. Table~\ref{tab:var_imp} lists the average variable importance of each life-space variable.
\begin{table}[ht]
\centering
\caption{Average importance of each life-space variable.} 
\label{tab:var_imp}
\begin{tabular}{lr}
  \hline\hline
Variable & Average Importance \\ 
  \midrule
  Errand trips (weekday) & 64.86 \\ 
  Home trips (weekday) & 21.54 \\ 
  Medical trips (weekday) & 6.75 \\ 
  Unknown trips (weekday) & 3.01 \\ 
  Social trips (weekday) & 1.86 \\ 
  Errand trips (weekend) & 0.87 \\ 
  Home trips (weekend) & 0.77 \\ 
  Unknown trips (weekend) & 0.24 \\ 
  Social trips (weekend) & 0.06 \\ 
  Medical trips (weekend) & 0.01 \\ 
  Work trips (weekday) & 0.00 \\ 
  work trips (weekend) & 0.00 \\ 
   \bottomrule
\end{tabular}
\end{table}

Among the life-space variables, ``Errand trips (weekday)'' stands out as the most important predictor, with a high average importance score of 64.86, suggesting that this variable plays a critical role in distinguishing between individuals with and without cognitive impairment. In contrast, ``Home trips (weekday)'' shows moderate importance with a score of 21.54, indicating a somewhat significant, but lesser role in the prediction model. Other variables, such as ``Medical trips (weekday)'' (6.75) and ``Unknown trips (weekday)'' (3.01), contribute minimally. The least important variables, such as ``Social trips (weekend)'' (0.06), ``Medical trips (weekend)'' (0.01), and both ``Work trips'' (weekday and weekend) with scores of 0.00, suggesting that work-related travel and medical trips on the weekend offer almost to no predictive value. This distribution highlights that weekend errand trips may serve as a key behavioral marker for cognitive impairment, while other trip types, particularly during weekends, contribute negligibly to the model's predictions.

\section{Conclusions}
We conducted a detailed analysis to understand how naturalistic driving behavior affects drivers with Mild Cognitive Impairment (MCI) or AD. To achieve this, we computed twelve life-space variables using geohashing methods, which captured the driving habits of the participants in our study. We then performed a two-fold analysis—both model-free and model-based—to assess the significance of these life-space variables in predicting MCI or AD.

 Our analysis revealed some intriguing findings. The frequency of errand trips emerged as the strongest predictor of drivers with MCI or AD, while work trips showed the least predictive power. Additionally, medical trips were identified as another crucial variable, as drivers with MCI or AD often have more frequent visits to hospitals or pharmacies compared to cognitively unimpaired drivers. Finally, our results indicated that drivers with MCI or AD are more likely to take trips to places outside their regular destinations, such as home and work, compared to those who are cognitively unimpaired. 

 It is important to understand that studies about driving behavior among older adults raise important ethical considerations, especially since there exists a potential for stigmatization or unintended consequences. Hence, care must be taken to avoid reinforcement of ageist assumptions or infringing on personal freedoms. Any data collected for such a study should be properly anonymized, have secure storage, and have transparent data sharing policies. Furthermore, any predictive models developed from such data should be used to assist/support, not replace human judgment in clinical decision making.

Looking ahead, our future research will focus on exploring methods to predict MCI or AD more accurately in drivers. This could be achieved by examining additional driving characteristics, such as speed, acceleration, the number of right turns, and the number of stops, and analyzing how effective these factors are in predicting MCI or AD alongside life-space variables. Additionally, our study cohort was exclusively focused on Nebraska, which, compared to many other states, exhibits low elevation variances. Expanding the study cohort into states with different elevation variances can lead to deeper insights from the life space variables.

\section*{Acknowledgements}
This work was funded by the National Institutes of Health (NIH), the National Institute on Aging (NIA R01AG17177), and the University of Nebraska Medical Center Mind \& Brain Health Labs. The views expressed in this paper are of the authors alone and not those of NIH or NIA. We thank our entire research team for coordinating this project. Because of personally identifying information (PII), the data is not available online. The authors declare no potential conflict of interest.

%\appendix

%\section{Appendix title 1}

%\section{Appendix title 2}

\sloppy
\bibliographystyle{elsarticle-harv} 
\bibliography{example}



%% For example:

%% \bibitem[Aladro et al.(2015)]{Aladro15} Aladro, R., Martín, S., Riquelme, D., et al. 2015, \aas, 579, A101

%%\end{thebibliography}

\end{document}